\documentclass[10pt,twocolumn,letterpaper]{article}

\usepackage[pagenumbers]{cvpr} 

\usepackage{graphicx}
\usepackage{amsmath}
\usepackage{amssymb}
\usepackage{booktabs}
\usepackage{multirow}
\usepackage{makecell}
\usepackage{booktabs}
\usepackage{colortbl}
\usepackage{xcolor}

\usepackage[pagebackref,breaklinks,colorlinks]{hyperref}

\usepackage[capitalize]{cleveref}
\crefname{section}{Sec.}{Secs.}
\Crefname{section}{Section}{Sections}
\Crefname{table}{Table}{Tables}
\crefname{table}{Tab.}{Tabs.}

\begin{document}

\title{Focal and Global Knowledge Distillation for Detectors}

\author{Zhendong Yang\thanks{This work was performed while Zhendong worked as an intern at ByteDance.}$^{~1,2}$\quad Zhe Li$^{2}$\quad Xiaohu Jiang$^{1}$\quad Yuan Gong$^{1}$\\ \quad Zehuan Yuan$^{2}$\quad  Danpei Zhao$^{3}$\quad Chun Yuan\thanks{Corresponding author}$^{~1}$\\
$^{1}$Tsinghua Shenzhen International Graduate School\quad $^{2}$ByteDance Inc\\$^{3}$BeiHang University\\
{\tt\small \{yangzd21,jiangxh21,gong-y21\}@mails.tsinghua.edu.cn \quad \{lizhe.axel,yuanzehuan\}@bytedance.com}\\
{\tt\small zhaodanpei@buaa.edu.cn \quad yuanc@sz.tsinghua.edu.cn}
}
\maketitle

\begin{abstract}
  Knowledge distillation has been applied to image classification successfully. However, object detection is much more sophisticated and most knowledge distillation methods have failed on it. In this paper, we point out that in object detection, the features of the teacher and student vary greatly in different areas, especially in the foreground and background. If we distill them equally, the uneven differences between feature maps will negatively affect the distillation. Thus, we propose Focal and Global Distillation (FGD). Focal distillation separates the foreground and background, forcing the student to focus on the teacher's critical pixels and channels. Global distillation rebuilds the relation between different pixels and transfers it from teachers to students, compensating for missing global information in focal distillation. As our method only needs to calculate the loss on the feature map, FGD can be applied to various detectors. We experiment on various detectors with different backbones and the results show that the student detector achieves excellent mAP improvement. For example, ResNet-50 based RetinaNet, Faster RCNN, RepPoints and Mask RCNN with our distillation method achieve 40.7\%, 42.0\%, 42.0\% and 42.1\% mAP on COCO2017, which are 3.3, 3.6, 3.4 and 2.9 higher than the baseline, respectively. Our codes are available at \url{https://github.com/yzd-v/FGD}.
\end{abstract}

\section{Introduction}
\label{sec:intro}

Recently, deep learning has achieved great success in various domains\cite{he2016deep,ren2015faster,ronneberger2015u,he2017mask}. To get better performance, we usually use a larger backbone, which needs more compute resources and inferences more slowly. To get over this,  knowledge distillation has been proposed\cite{hinton2015distilling}. Knowledge distillation is a method to inherit the information from a large teacher network to a compact student network and achieve strong performance without extra cost during inference time. However, most distillation methods\cite{zagoruyko2016paying, yim2017gift, heo2019comprehensive, tung2019similarity} are designed for image classification, which lead to trivial improvements for object detection.

\begin{figure}
  \centering
  \includegraphics[width=0.9\linewidth]{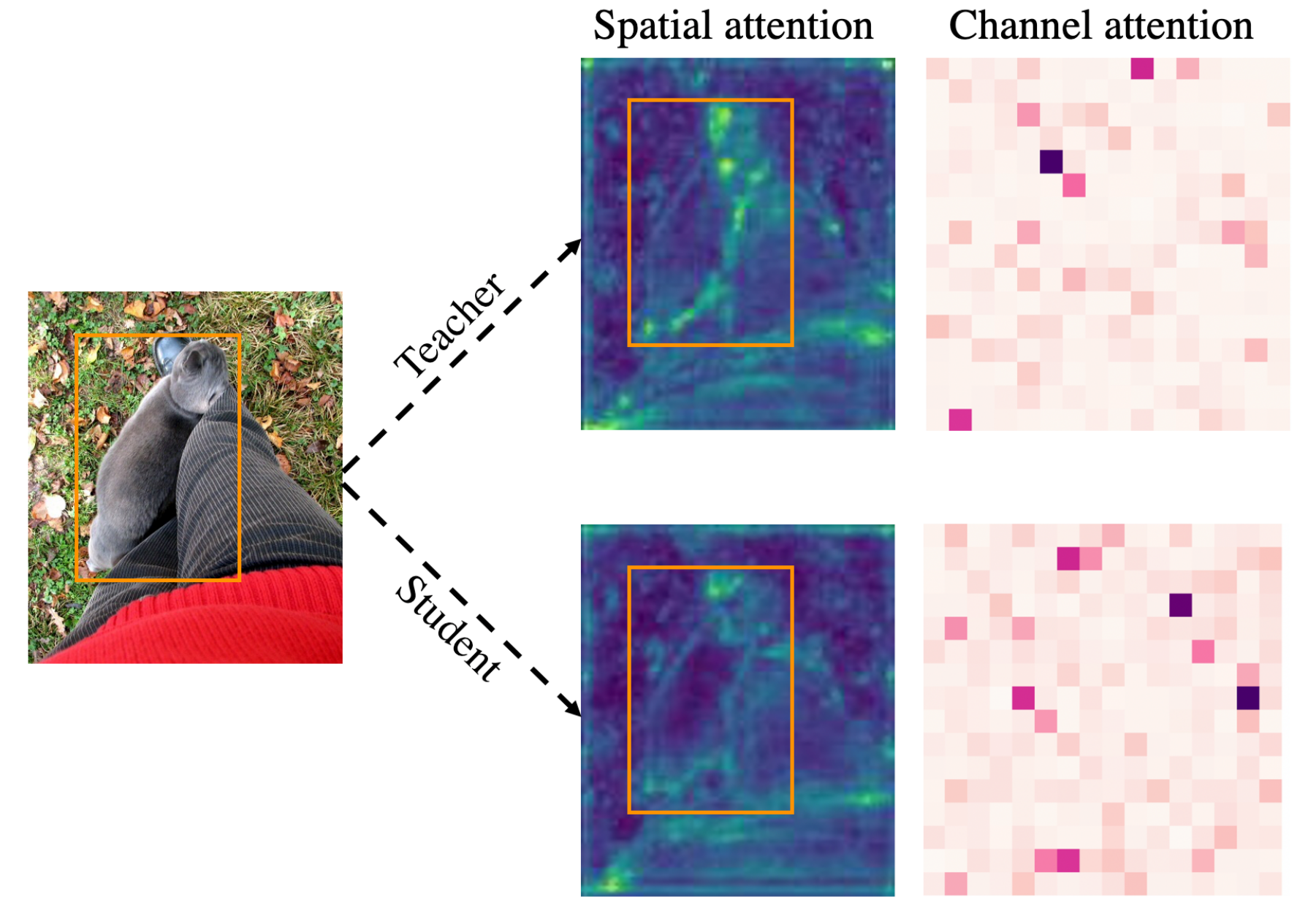}
  \caption{Visualization of the spatial and channel attention map from the teacher detector (RetinaNet-ResNeXt101) and the student detector (RetinaNet-ResNet50).}
  \label{figure:heat map}
\end{figure}

\begin{table}
  \centering
  \begin{tabular}{@{}l|ccc|cc}
    \toprule
    \multirow{7}{*}{\makecell[c]{RetinaNet\\Res101-Res50}} & 
    \multicolumn{3}{c|}{distillation area}&
    \multirow{2}{*}{mAP} &
    \multirow{2}{*}{mAR}\\
    \cmidrule{2-4}
    &fg &bg&split &\multicolumn{2}{c}{}\\
    \cmidrule{2-6}
    &$\times$&$\times$&&37.4&53.9\\
    &$\checkmark$ &$\times$& &39.3&55.6\\
    &$\times$&$\checkmark$&&39.2&55.8\\
    &$\checkmark$ &$\checkmark$&$\times$&38.9&55.1\\
    &$\checkmark$ &$\checkmark$&$\checkmark$&{\bf39.4}&{\bf56.1}\\
    \bottomrule
  \end{tabular}
  \caption{Comparisons of different distillation areas. {\bf fg}: foreground. {\bf bg}: background. {\bf split}: split the foreground and background and distill them with different weights.}
  \label{table:fbg ablation}
\end{table}

It is well acknowledged that the extreme foreground-background class imbalance is a key point in object detection\cite{lin2017focal}. The imbalanced ratio also harms the distillation for object detection. There are some efforts for this problem. Chen \etal~\cite{chen2017learning} distributes a weight to suppress the background. Mimick\cite{li2017mimicking} distills the positive area proposed by region proposal network of the student. FGFI\cite{wang2019distilling} and TADF\cite{sun2020distilling} use the fine-grained and Gaussian Mask to select the distillation area, respectively. Defeat\cite{guo2021distilling} distills the foreground and background separately. However, where is the key area for distillation is still not clear.

In order to explore the difference between the features of students and teachers, we do the visualization of the spatial and channel attention. As the \cref{figure:heat map} shows, the difference between student's attention and teacher's attention in the foreground is quite significant, while that in the background is relatively small. This may lead to different difficulties in learning the foreground and background. In this paper, we further explore the influence of the foreground and background in knowledge distillation on object detection. We design experiments by decoupling the foreground and background in the distillation. Surprisingly, as shown in \cref{table:fbg ablation}, the performance of distillation on the foreground and background together is the worst, even worse than only using foreground or background.  This phenomenon suggests that the uneven differences in the feature map can negatively affect distillation. Besides, as shown in \cref{figure:heat map}, the attention between each channel is also very different. Thinking one step deeper, not only are there negative influences between the foreground and the background, but also between the pixels and the channels. Therefore, we propose focal distillation. While separating the foreground and background, focal distillation also calculates the attention of different pixels and channels in teacher's feature, allowing the student to focus on teacher's crucial pixels and channels.

However, just focusing on key information is not enough. It is well known that global context also plays an important role in detection. A lot of relation modules have been successfully applied into detection, such as non-local\cite{wang2018non}, GcBlock\cite{cao2019gcnet}, relation network\cite{hu2018relation}, which have greatly improved the performance of detectors. In order to compensate for the missing global information in focal distillation, we further propose global distillation. In global distillation, we utilize GcBlock to extract the relation between different pixels and then distill them from teachers to students.

\begin{figure*}
  \centering
  \includegraphics[width=0.97\linewidth]{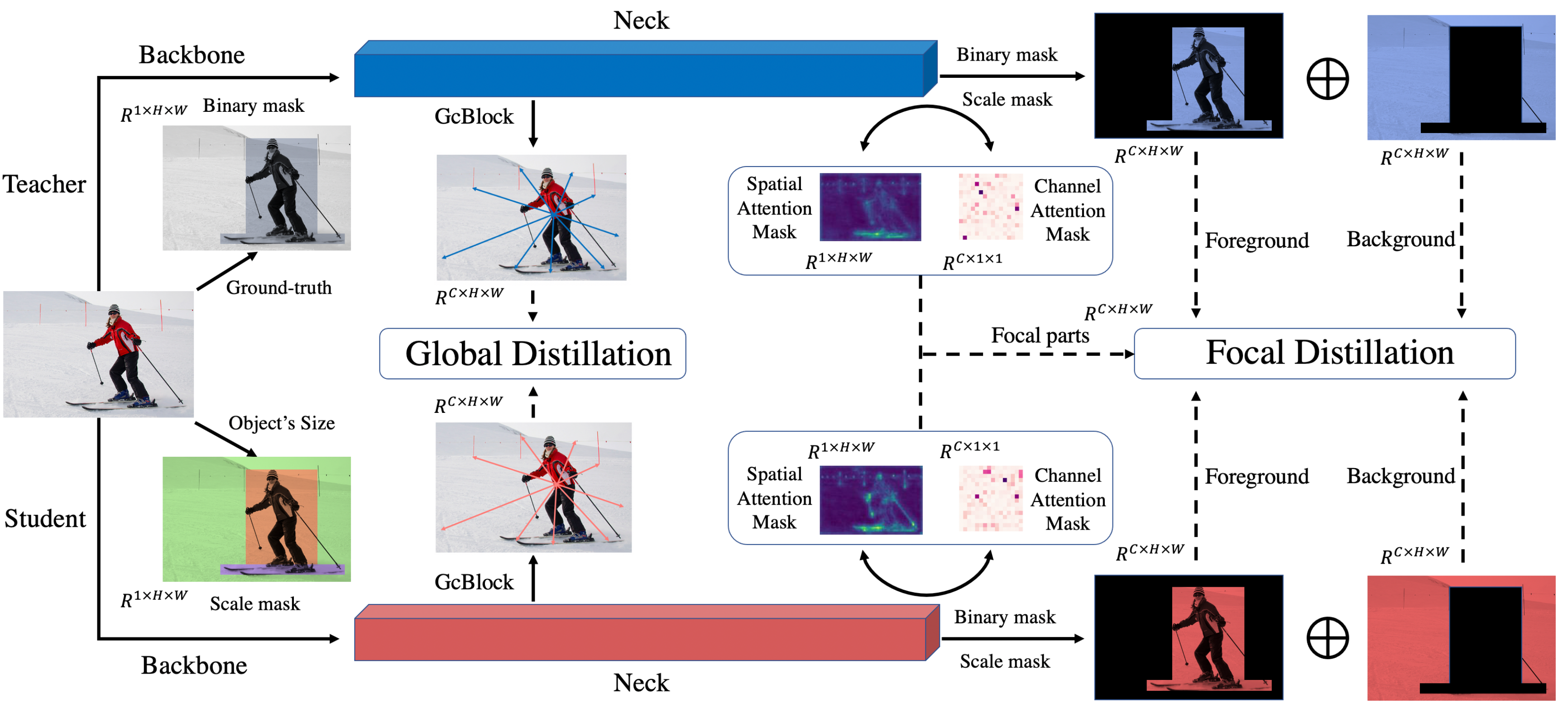}
  \caption{An illustration of FGD, including focal distillation and global distillation. Focal distillation not only separates the foreground and the background, but also enables the student network to better pay attention to the important information in the teacher network's feature map. Global distillation bridges the gap between the global context of the student and the teacher.}
  \label{figure:structure}
\end{figure*}

As we analyzed above, we propose \textbf{F}ocal and \textbf{G}lobal \textbf{D}istillation (FGD), combining focal distillation and global distillation, as shown in  \cref{figure:structure}. All loss functions are only calculated on features, so that FGD can be used directly on various detectors, including two-stage models, anchor-based one-stage models and anchor-free one-stage models. Without bells and whistles, we achieve state-of-the-art performances in object detection with FGD. In a nutshell, the contributions of this paper are:

\begin{itemize}
  \item
  We present that the pixels and channels that teacher and student pay attention to are quite different. If we distill the pixels and channels without distinguishing them, it will result in a trivial improvement.
  \item
  We propose focal and global distillation, which enables the student not only to focus on the teacher's critical pixels and channels, but also to learn the relation between pixels.
  \item
  We verify the effectiveness of our method on various detectors via extensive experiments on the COCO\cite{lin2014microsoft}, including one-stage, two-stage, anchor-free methods, achieving state-of-the-art performance.
\end{itemize}

\section{Related Work}
\label{sec:related work}

\subsection{Object Detection}

Object detection is a fundamental and challenging task in computer vision. The CNN-based detection networks with high performance are divided into two-stage\cite{ren2015faster,he2017mask,cai2018cascade}, anchor-based one-stage\cite{liu2016ssd,lin2017focal,redmon2018yolov3} and anchor-free one-stage detectors\cite{tian2019fcos,yang2019reppoints,duan2019centernet}. One-stage detectors get the classification and bounding box of targets on feature maps directly. In contrast, two-stage detectors utilize RPN and RCNN head to achieve better results but cost more time. Prior anchor boxes provide one-stage models with proposals to detect targets. However, the number of anchor boxes is far more than targets, which brings extra computation. While anchor-free detectors show a way to predict the key point and location of targets directly. Although there are different detection heads, their inputs are all features. Therefore, our feature-based knowledge distillation method can be applied in almost all detectors.

\subsection{Knowledge Distillation}
Knowledge distillation is a method of model compression without changing the network structure. It is first proposed by Hinton \etal~\cite{hinton2015distilling}, which uses the output as soft labels to transfer the dark knowledge from a large teacher network to a small student network for the classification task. Moreover, FitNet\cite{romero2014fitnets} proves that the semantic information from intermediate is also helpful to guide the student model. There have been many works\cite{zagoruyko2016paying, yim2017gift, heo2019comprehensive, tung2019similarity} that improve the student classifiers significantly.

Recently, some works have successfully applied knowledge distillation to detectors. Chen \etal~\cite{chen2017learning} first apply knowledge distillation to detection by distilling knowledge on the neck feature, the classification head, and the regression head. Nevertheless, distilling the whole feature may introduce much noise because of the imbalance between the foreground and background. Li \etal~\cite{li2017mimicking} choose the features sampled from RPN to calculate distillation loss. Wang \etal~\cite{wang2019distilling} propose the fine-grained mask to distill the regions calculated by ground-truth bounding boxes. Sun \etal~\cite{sun2020distilling} utilize the Gaussian Mask to cover the ground-truth for distillation. Such methods lack the distillation for the background. Without distinguishing the foreground and background, GID\cite{dai2021general} distills the areas where the performance of the student and teacher is different. Guo \etal~\cite{guo2021distilling} shows that both the foreground and background play important roles for distillation, and distilling them separately benefits the student more. Both of their methods distill the knowledge from the background and get significant results. However, they treat all the pixels and channels equally. FKD\cite{zhang2020improve} uses attention masks and Non-local module\cite{wang2018non} to guide the student and distills the relation, respectively. However, it distills the foreground and background together.

The critical problem of distillation for detection is to select the valuable area for distillation. The previous distillation methods treat all the pixels and channels equally\cite{wang2019distilling,sun2020distilling,dai2021general,guo2021distilling} or distill all the areas\cite{zhang2020improve} together. Most methods lack the distillation of the global context information. In this paper, we use ground-truth boxes to separate the images, and then use the attention masks from the teacher to select crucial parts for distillation. In addition, we capture the global relations between different pixels and distill them to the student, which brings another improvement.

\section{Method}
\label{sec:method}

Most detectors have used FPN\cite{lin2017feature} to utilize the multi-scale semantic information. The features from FPN fuse different levels of semantic information from the backbone and are used to predict directly. Transferring the knowledge of these features from the teacher has significantly improved the performance of the student. Generally, the distillation of the features can be formulated as:
\begin{equation}
    L_{fea}=\frac{1}{CHW}\sum_{k=1}^{C}\sum_{i=1}^{H}\sum_{j=1}^{W}\big( F_{k,i,j}^{T}-f(F_{k,i,j}^{S})\big)^{2}
  \label{general_feature_loss}
\end{equation}
where $F^{T}$ and $F^{S}$ denote the feature from the teacher and student, respectively, and $f$ is the adaptation layer to reshape the $F^{S}$ to the same dimension as $F^{T}$. $H, W$ denote the height and width of the feature and $C$ is the channel.

However, such methods treat all the parts equally and lack the distillation of the global relations between different pixels. To get over the above problems, we propose FGD, which includes focal and global distillation, as shown in \cref{figure:structure}. Here we will introduce our method in detail.

\subsection{Focal Distillation}
\label{sec:focal dis}
\begin{figure}
  \centering
  \includegraphics[width=0.97\linewidth]{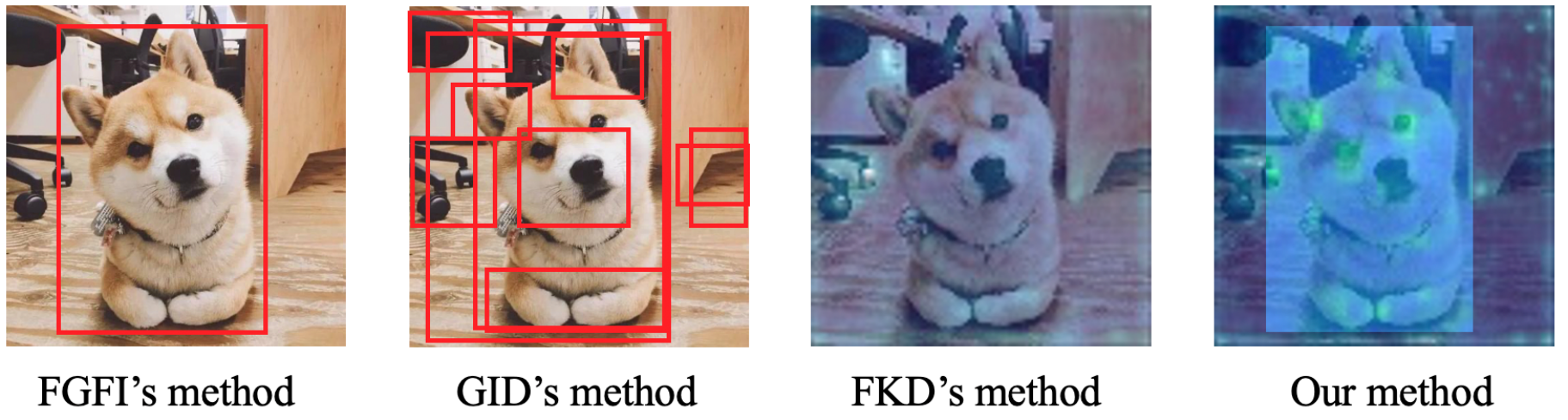}
  \caption{Comparison of the distillation areas between our method (FGD) and other methods. FGFI and GID only distill the areas in the red bounding box. The areas where GID and FKD distill are changeable during training. Different colors mean different weights and the green parts mean the spatial attention pixels.}
  \label{figure:selective area}
\end{figure}

For the foreground and background imbalance, we propose focal distillation to separate the images and guide the student to focus on crucial pixels and channels. The comparison of the distillation areas can be seen in \cref{figure:selective area}.

Firstly we set a binary mask $M$ to separate the background and foreground:
\begin{equation}
    \label{mask}
    M_{i,j}=
    \begin{cases}
        1, & \text{if}\ \ (i,j)\in r \\ 
        0, & \text{Otherwise}
    \end{cases}
\end{equation}
where $r$ denotes the ground-truth boxes and $i, j$ are the the horizontal and vertical coordinates of the feature map, respectively. If $(i, j)$ falls in the ground truth, then $M_{i,j} = 1$, otherwise it is 0.

The targets with larger-scale will occupy more loss because they own more pixels, which will influence the distillation of the small targets. And the ratios of foreground to background vary greatly in different images. Therefore, in order to treat different targets equally and balance the loss of foreground and background, we set a scale mask $S$ as:
\begin{equation}
    \label{s-mask}
    S_{i,j}=
    \begin{cases}
        \frac{1}{H_{r}W_{r}}, & \text{if}\ \ (i,j)\in r \\ 
        \frac{1}{N_{bg}}, & \text{Otherwise}
    \end{cases}
\end{equation}
\begin{equation}
    \label{Nbg}
    N_{bg}=\sum_{i=1}^{H}\sum_{j=1}^{W}(1- M_{i,j})
\end{equation}
where $H_{r}$ and $W_{r}$ denote the height and width of the ground-truth box $r$. If a pixel belongs to different targets, we choose the smallest box to calculate the $S$.

SENet\cite{hu2018squeeze} and CBAM\cite{woo2018cbam} show that focusing on crucial pixels and channels helps CNN-based models get better results. Zagoruyko \etal~\cite{zagoruyko2016paying} use a simple way to get the spatial attention mask and improve the performance of distillation. In this paper, we apply a similar method to select focal pixels and channels, and then get corresponding attention masks. We calculate the absolute mean values on different pixels and different channels, respectively:

\begin{equation}
    \label{ms-a}
    G^{S}(F)=\frac{1}{C}\cdot\sum_{c=1}^{C}\left | F_{c}\right |
\end{equation}

\begin{equation}
    \label{mc-a}
    G^{C}(F)=\frac{1}{HW}\cdot\sum_{i=1}^{H}\sum_{j=1}^{W}|F_{i,j}|
\end{equation}
where $H$, $W$, $C$ denote the feature's height, width, and channel. $G^{S}$ and $G^{C}$ are the spatial and channel attention map. Then the attention mask can be formulated as:

\begin{equation}
    \label{s-a}
    A^{S}(F)=H\cdot W\cdot softmax\big(G^{S}(F)/T\big)
\end{equation}
\begin{equation}
    \label{c-a}
    A^{C}(F)=C\cdot softmax\big(G^{C}(F)/T\big)
\end{equation}
where $T$ is the temperature hyper-parameter proposed by Hinton \etal~\cite{hinton2015distilling} to adjust the distribution.

There are significant differences between the masks of the student and teacher. In training process, we use the teacher's masks to guide the student. With the binary mask $M$, scale mask $S$, attention mask $A^{S}$ and $A^{C}$, we propose the feature loss $L_{fea}$ as follows:

\begin{equation}
    \label{l-decoupled}
    \begin{split}
        L_{fea}=\alpha\sum_{k=1}^{C}\sum_{i=1}^{H}\sum_{j=1}^{W} M_{i,j} S_{i,j} A_{i,j}^{S} A_{k}^{C}\big( F_{k,i,j}^{T}-f(F_{k,i,j}^{S})\big)^{2} \\
        + \beta\sum_{k=1}^{C}\sum_{i=1}^{H}\sum_{j=1}^{W} (1-M_{i,j}) S_{i,j} A_{i,j}^{S} A_{k}^{C}\big( F_{k,i,j}^{T}-f(F_{k,i,j}^{S})\big)^{2}
    \end{split}
\end{equation}
where $A^{S}$ and $A^{C}$ denote the spatial and channel attention mask of the teacher detector, respectively. $F^{T}$ and $F^{S}$ denote the feature maps of the teacher detector and student detector, respectively. $\alpha$ and $\beta$ are the hyper-parameters to balance the loss between foreground and background.

Besides, we use attention loss $L_{at}$ to force the student detector to mimic the spatial and channel attention mask of the teacher detector, which is formulated as:

\begin{equation}
    \label{l-attention}
        L_{at}=\gamma\cdot \big(l(A_{t}^{S}, A_{S}^{S}) + l(A_{t}^{C}, A_{S}^{C})\big)
\end{equation}
where $t$ and $s$ denote the teacher and student. $l$ denotes L1 loss and $\gamma$ is a  hyper-parameter to balance the loss.

The focal loss $L_{focal}$ is the sum of feature loss $L_{at}$ and attention loss $L_{at}$:
\begin{equation}
    \label{l-focal}
        L_{focal}= L_{fea} +  L_{at}
\end{equation}

\subsection{Global Distillation}
The relation\cite{wang2018non, hu2018relation, cao2019gcnet} between different pixels has valuable knowledge and is utilized to improve the performance for detection tasks. And in \cref{sec:focal dis}, we utilize Focal Distillation to separate the images and force the student focus on crucial parts. However, such distillation cuts off the relation between foreground and background. So here we propose Global Distillation, which aims to extract the global relation between different pixels from the feature maps and distill it from the teacher to the student.

\begin{figure}
  \centering
  \includegraphics[width=1.0\linewidth]{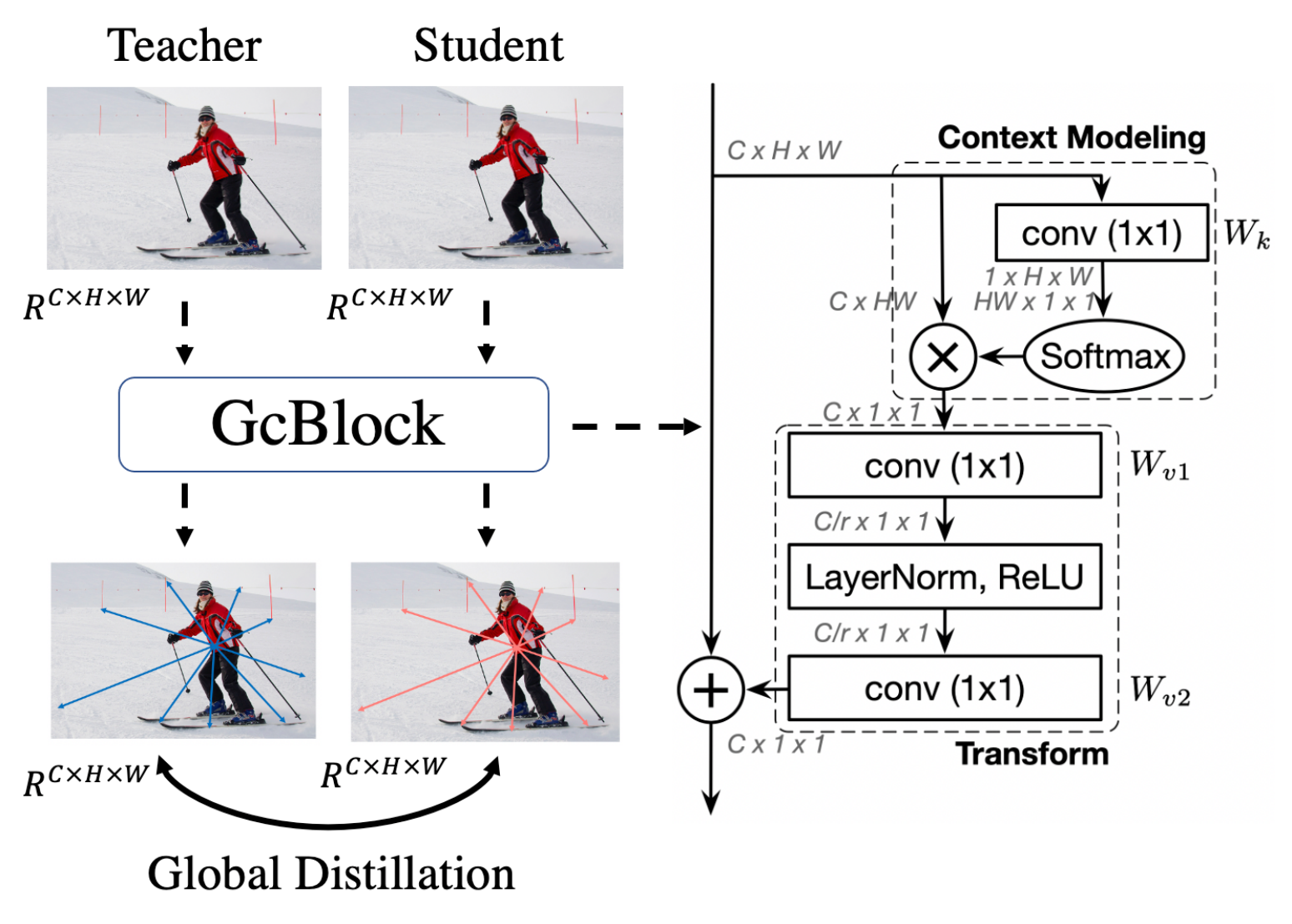}
  \caption{The Global Distillation with GcBlock. The inputs are the feature maps from the teacher's neck and student's neck, respectively.}
  \label{figure:gcblock}
\end{figure}

As shown in \cref{figure:gcblock}, we utilize GcBlock\cite{cao2019gcnet} to capture the global relation information in a single image and force the student detector to learn the relation from the teacher detector. The global loss $L_{global}$ is as follows:
\begin{equation}
\begin{split}
    \label{l-global}
        L_{global}=&\lambda\cdot\sum\Big(\mathcal{R}\big( F^{T}\big) - \mathcal{R}\big(F^{S}\big)\Big)^{2} \\
        \mathcal{R}(F)=&F + W_{v2}(ReLU(LN(W_{v1} \\
        &(\sum_{j=1}^{N_{p}}\frac{e^{W_{k}F_{j}}}{\sum_{m=1}^{N_{p}}e^{W_{k}F_{M}}}F_{j}))))
\end{split}
\end{equation}
where $W_{k}$, $W_{v1}$ and $W_{v2}$ denote convolutional layers, $LN$ denotes the layer normalization, $N_{p}$ is the number of pixels in the feature and $\lambda$ is a hyper-parameter to balance the loss.

\subsection{Overall loss}
To sum up, we train the student detector with the total loss as follows:
\begin{equation}
    \label{l-all}
        L=L_{original}+L_{focal}+L_{global}
\end{equation}
where $L_{original}$ is the original loss for detectors.

The distillation loss is calculated just on feature maps, which can be obtained from the neck of the detectors. So it can be easily applied to different detectors.

\section{Experiments}
\label{sec:experiments}
\subsection{Dataset}
We evaluate our knowledge distillation method on COCO dataset\cite{lin2014microsoft}, which contains 80 object classes. We use the 120k train images for training and 5k val images for testing for all the experiments. The performances of different detectors are evaluated in Average Precision and Average Recall.

\subsection{Details}
We conduct experiments on different detection frameworks, including two-stage models\cite{ren2015faster}, anchor-based one-stage models\cite{lin2017focal}, and anchor-free one-stage models\cite{yang2019reppoints,tian2019fcos}. Besides, we verify our method on the Mask RCNN\cite{he2017mask} and get significant improvement for instance segmentation. Kang \etal~\cite{kang2021instance} propose inheriting strategy which initializes the student with the teacher's neck and head parameters and gets better results. Here we use this strategy to initialize the student which has the same head structure as the teacher. All the experiments are conducted with mmdetection\cite{mmdetection} with Pytorch\cite{paszke2019pytorch}.

FGD uses $\alpha, \beta, \gamma, \lambda$ to balance the loss of foreground and background in \cref{l-decoupled}, attention loss in \cref{l-attention} and global loss in \cref{l-global}, respectively. And $T=0.5$ is used to adjust the attention distribution for all the experiments. We adopt the hyper-parameters$\{\alpha = 5 \times 10^{-5}, \beta=2.5\times10^{-5}, \gamma=5\times10^{-5}, \lambda=5\times10^{-7}\}$ for all the two-stage models, $\{\alpha = 1 \times 10^{-3}, \beta=5\times10^{-4}, \gamma=1\times10^{-3}, \lambda=5\times10^{-6}\}$ for all the anchor-based one-stage models, $\{\alpha = 1.6 \times 10^{-3}, \beta=8\times10^{-4}, \gamma=8\times10^{-3}, \lambda=8\times10^{-6}\}$ for all the anchor-free one-stage models. We train all the detectors for 24 epochs with SGD optimizer, which the momentum is 0.9 and the weight decay is 0.0001.

\begin{table}
  \centering
  \begin{tabular}{l|cccc}
    \toprule
    Method & mAP & AP$_{S}$ & AP$_{M}$ & AP$_{L}$\\
    \midrule
    RetinaNet-Res101(T) & 38.9 &21.0&42.8&52.4\\
    RetinaNet-Res50(S) & 37.4 &20.6&40.7&49.7\\
    FGFI\cite{wang2019distilling}& 38.6&21.4&42.5&51.5\\
    GID\cite{dai2021general} & 39.1&22.8&43.1&52.3\\
    \rowcolor{lightgray!45}Ours & 39.6&22.9&43.7&53.6 \\
    \rowcolor{lightgray!45}Ours \dag & 39.7&22.0&43.7&53.6\\
    \midrule
    RCNN-Res101(T) & 39.8 &22.5&43.6&52.8\\
    RCNN-Res50(S) & 38.4 &21.5&42.1&50.3\\
    FGFI\cite{wang2019distilling}& 39.3&22.5&42.3&52.2\\
    GID\cite{dai2021general} & 40.2&22.7&44.0&53.2\\
    \rowcolor{lightgray!45}Ours & 40.4&22.8&44.5&53.5\\
    \rowcolor{lightgray!45}Ours \dag & 40.5&22.6&44.7&53.2\\
    \midrule
    FCOS-Res101(T) & 40.8 &24.2&44.3&52.4\\
    FCOS-Res50(S) & 38.5 &21.9&42.8&48.6\\
    GID\cite{dai2021general} & 42.0&25.6&45.8&54.2\\
    \rowcolor{lightgray!45}Ours & 42.1&27.0&46.0&54.6\\
    \rowcolor{lightgray!45}Ours \dag & 42.7&27.2&46.5&55.5\\
    \bottomrule
  \end{tabular}
  \caption{Results of different distillation methods with different detection frameworks on COCO dataset. {\bf T} and {\bf S} mean the teacher and student detector, respectively. FGFI can only be applied to an anchor-based detector. \dag \ means using inheriting strategy. We train the FCOS with tricks including GIoULoss, norm-on-bbox and center-sampling which is the same as GID.}
  \label{table:main results}
\end{table}

\begin{table*}
  \centering
  \begin{tabular}{c|l|lccc|cccc}
    \toprule
    Teacher& Student & mAP  & AP$_{S}$ & AP$_{M}$ &AP$_{L}$&mAR& AR$_{S}$ & AR$_{M}$ &AR$_{L}$\\
    \midrule
    \multirow{4}{*}{\makecell{RetinaNet\\ResNeXt101}}
    &RetinaNet-Res50 & 37.4 &20.6&40.7&49.7&53.9&33.1&57.7&70.2\\
    &FKD\cite{zhang2020improve} & 39.6(+2.2)&22.7&43.3&52.5&56.1(+2.2)&36.8&60.0&72.1\\
    &\cellcolor{lightgray!45}Ours & \cellcolor{lightgray!45}40.4(+3.0)&\cellcolor{lightgray!45}23.4&\cellcolor{lightgray!45}44.7&\cellcolor{lightgray!45}54.1&\cellcolor{lightgray!45}56.7(+2.8)&\cellcolor{lightgray!45}37.6&\cellcolor{lightgray!45}61.5&\cellcolor{lightgray!45}72.4\\
    &\cellcolor{lightgray!45}Ours\dag & \cellcolor{lightgray!45}40.7(+3.3)&\cellcolor{lightgray!45}22.9&\cellcolor{lightgray!45}45.0&\cellcolor{lightgray!45}54.7&\cellcolor{lightgray!45}56.8(+2.9)&\cellcolor{lightgray!45}36.5&\cellcolor{lightgray!45}61.4&\cellcolor{lightgray!45}72.8\\
    \midrule
    \multirow{3}{*}{\makecell{Cascade\\Mask RCNN\\ResNeXt101}}
    &Faster RCNN-Res50 & 38.4 &21.5&42.1&50.3&52.0&32.6&55.8&66.1\\
    &FKD\cite{zhang2020improve} & 41.5(+3.1)&23.5&45.0&55.3&54.4(+2.4)&34.0&58.2&69.9\\
    &\cellcolor{lightgray!45}Ours  &\cellcolor{lightgray!45}42.0(+3.6)&\cellcolor{lightgray!45}23.8&\cellcolor{lightgray!45}46.4&\cellcolor{lightgray!45}55.5&\cellcolor{lightgray!45}55.4(+3.4)&\cellcolor{lightgray!45}35.5&\cellcolor{lightgray!45}60.0&\cellcolor{lightgray!45}70.0\\
    \midrule
    \multirow{4}{*}{\makecell{RepPoints\\ResNeXt101}}
    &RepPoints-Res50 & 38.6&22.5&42.2&50.4&55.1&34.9&59.4&70.3\\
    &FKD\cite{zhang2020improve} & 40.6(+2.0)&23.4&44.6&53.0&56.9(+1.8)&37.3&60.9&71.4\\
    &\cellcolor{lightgray!45}Ours & \cellcolor{lightgray!45}41.3(+2.7)&\cellcolor{lightgray!45}24.5&\cellcolor{lightgray!45}45.2&\cellcolor{lightgray!45}54.0&\cellcolor{lightgray!45}58.4(+3.3)&\cellcolor{lightgray!45}39.1&\cellcolor{lightgray!45}62.9&\cellcolor{lightgray!45}74.2\\
    &\cellcolor{lightgray!45}Ours\dag & \cellcolor{lightgray!45}42.0(+3.4)&\cellcolor{lightgray!45}24.0&\cellcolor{lightgray!45}45.7&\cellcolor{lightgray!45}55.6&\cellcolor{lightgray!45}58.2(+3.1)&\cellcolor{lightgray!45}37.8&\cellcolor{lightgray!45}62.2&\cellcolor{lightgray!45}73.3\\
    \midrule
    \multirow{2}{*}{Teacher} &
    \multirow{2}{*}{Student} &
    \multicolumn{4}{c|}{Boundingbox AP}&
    \multicolumn{4}{c}{Mask AP} \\
    \cmidrule{3-10}
    & &mAP &AP$_{S}$ & AP$_{M}$ & AP$_{L}$& mAP&  AP$_{S}$ &  AP$_{M}$ & AP$_{L}$\\
    \midrule
    \multirow{3}{*}{\makecell{Cascade\\Mask RCNN\\ResNeXt101}}
    &Mask RCNN-Res50 & 39.2&22.9&42.6&51.2&35.4&19.1&38.6&48.4\\
    &FKD\cite{zhang2020improve} & 41.7(+2.5)&23.4&45.3&55.8&37.4(+2.0)&19.7&40.5&52.1\\
    &\cellcolor{lightgray!45}Ours & \cellcolor{lightgray!45}42.1(+2.9)&\cellcolor{lightgray!45}23.7&\cellcolor{lightgray!45}46.2&\cellcolor{lightgray!45}55.7&\cellcolor{lightgray!45}37.8(+2.4)&\cellcolor{lightgray!45}19.7&\cellcolor{lightgray!45}41.3&\cellcolor{lightgray!45}52.3\\
    \bottomrule
  \end{tabular}
  \caption{Results of more detectors with stronger teacher detectors on COCO dataset. \dag\ means using inheriting strategy, which can only be applied when the student and teacher have the same head structure.}
  \label{table:more results}
\end{table*}

\subsection{Main Results}
Our method can be applied to different detection frameworks easily, so we first conduct experiments on three popular detectors, including a two-stage detector (Faster RCNN), an anchor-based one-stage detector (RetinaNet) and an anchor-free detector (FCOS). We compare with other two knowledge distillation methods\cite{wang2019distilling,dai2021general} for object detection. In the experiments, we choose the detectors with ResNet-50\cite{he2016deep} as the students and the identical detectors with ResNet-101 as the teachers. As shown in \cref{table:main results}, our distillation method surpasses the other two state-of-the-art methods. All the student detectors gain significant AP improvements with the knowledge transferred from teacher detectors, {\em e.g. }the RetinaNet based ResNet-50 gets 2.3 mAP improvement on COCO dataset. Furthermore, in this Res101-Res50 setting, the student detectors even outperform the teacher detectors by training with FGD. 

\subsection{Distillation of more detectors with stronger students and teachers}
Our method can also be applied between heterogeneous backbones, {\em e.g. }the ResNeXt\cite{xie2017aggregated} based teacher detector distill the ResNet based student detector. Here we conduct experiments on more detectors and use stronger backbone-based teacher detectors. And we compare the results with FKD\cite{wang2019distilling}, which is another effective and general distillation method. As shown in \cref{table:more results}, all the student detectors achieve significant improvements on both AP and AR.
Besides, comparing the results with \cref{table:main results}, we find that student detectors perform better with stronger teacher detectors, {\em e.g. }Retina-Res50 model achieves 40.7 and 39.7 mAP with ResNeXt101 and ResNet101 based teacher, respectively. The comparisons show that student detectors get the better feature by mimicking the feature maps of stronger backbones-based teacher detectors.

FGD only needs to calculate the distillation loss on the feature maps. So we also apply our method to Mask RCCN for object detection and instance segmentation. And in this experiment, we use the bounding box labels for the focal distillation. As shown in \cref{table:more results}, our method brings 2.9 Boundingbox AP gains and 2.4 Mask AP gains, which proves our distillation method is also effective for instance segmentation.

\begin{figure*}
  \centering
  \includegraphics[width=0.8\linewidth]{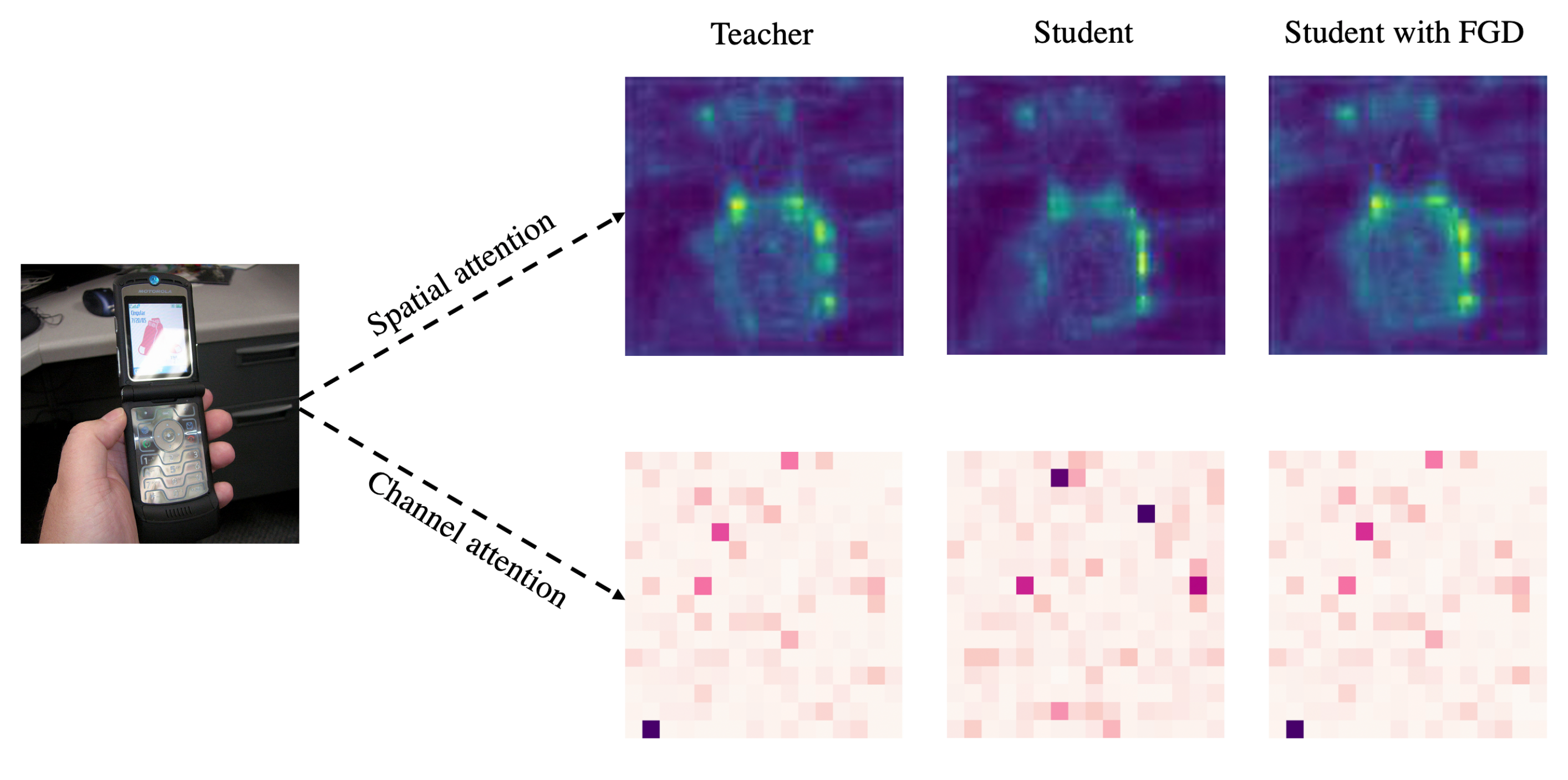}
  \caption{Visualization of the spatial and channel attention mask from different detectors. Each pixel in the channel attention mask means a channel. {\bf Teacher detector}: RetinaNet-ResNeXt101. {\bf Student detector}: RetinaNet-ResNet50}
  \label{figure:heat_map_compare_sc}
\end{figure*}

\subsection{Better feature with FGD}
As shown in \cref{table:main results} and \cref{table:more results}, initializing the student with the teacher's neck and head parameters brings another improvement, which indicates the student gets a similar feature with the teacher. So in this subsection, we visualize and compare the spatial attention mask and channel attention mask from teacher detector, student detector and student detector with FGD, which is shown in \cref{figure:heat_map_compare_sc}. Comparing the attention masks between the teacher and student, we can see they have a big difference in the distribution of pixels and channels before distillation, {\em e.g.} the teacher detector focuses more on the fingers and has a larger weight in channel 241. However, after training with FGD, the student detector has a similar distribution of pixels and channels with the teacher detectors, which means the student focuses on the same parts as the teacher. This also explains how FGD helps the student detector perform better. Based on a similar feature, the student detector gets significant improvements and even outperforms the teacher detector.

\subsection{Analysis}
\subsubsection{Sensitivity study of different losses}
In this paper, we transfer the focal knowledge and global knowledge from the teacher to the student. In this subsection, we conduct experiments of focal loss ($L_{focal}$) and global loss ($L_{global}$) to investigate their influences on the student with RetinaNet. As shown in \cref{table:ablation study}, both the focal loss and global loss lead to significant AP and AR improvements. Furthermore, considering targets with different sizes, we find $L_{focal}$ benefits more to the large size targets and $L_{global}$ benefits more to the small and medium targets. Besides, when combining $L_{focal}$ and $L_{global}$, we achieve 40.4 mAP and 56.7 mAR, which indicates the focal loss and global loss are complementary to each other.
\begin{table}
  \centering
  \begin{tabular}{@{}l|c|ccc}
    \toprule
    Method & \multicolumn{4}{c}{ReinaNet ResX101-Res50}\\
    \midrule
    L$_{focal}$  & - &\checkmark&-&\checkmark\\
    L$_{global}$ & - &-&\checkmark&\checkmark\\
    \midrule
    mAP & 37.4 &40.2&40.2&{\bf40.4}\\
    AP$_{S}$ & 20.0 & 22.8&22.9&{\bf23.4}\\
    AP$_{M}$ & 40.7 &44.0&44.3&{\bf44.7}\\
    AP$_{L}$ & 49.7 & 54.0&53.4&{\bf54.1}\\
    \midrule
    mAR & 53.9 &56.2&56.4&{\bf56.7}\\
    AR$_{S}$ & 33.1 &36.8&37.3&{\bf37.6}\\
    AR$_{M}$ & 57.7 &60.3&60.5&{\bf61.5}\\
    AR$_{L}$ & 70.2 &72.3&72.2&{\bf72.4}\\
    \bottomrule
  \end{tabular}
  \caption{Ablation study of focal and global distillation.}
  \label{table:ablation study}
\end{table}

\subsubsection{Sensitivity study of focal distillation}
 In focal distillation, we use the ground-truth boxes to separate the images and guide the student with the teacher's attention masks. In this subsection, we explore the effectiveness of focal distillation.
 
\begin{figure}
  \centering
  \begin{subfigure}{0.495\linewidth}
    \includegraphics[width=1\linewidth]{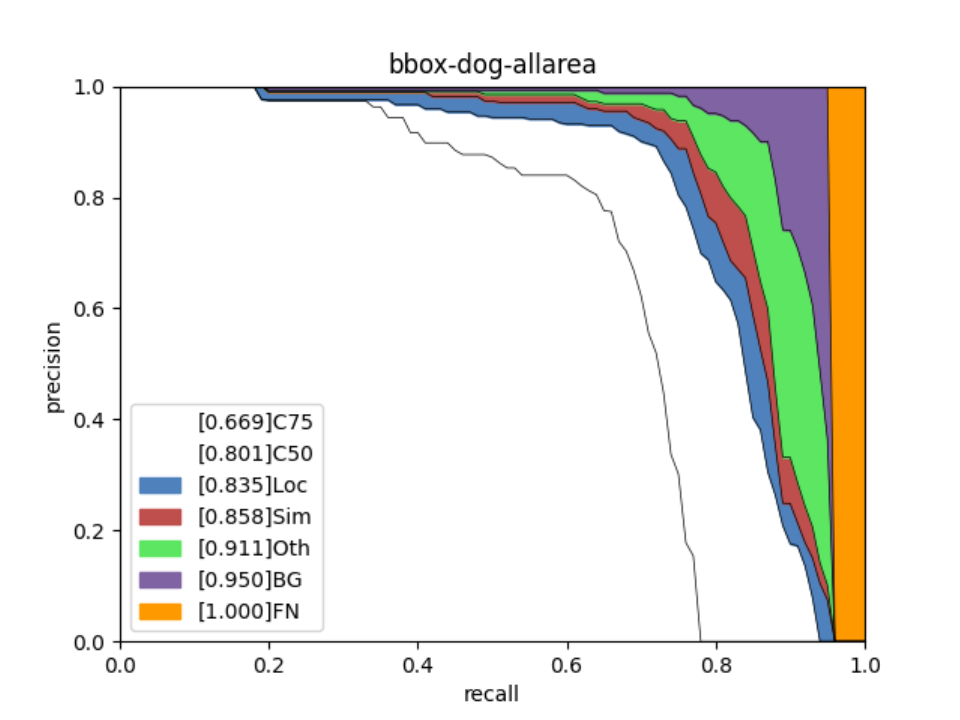}
    \caption{KD: foreground}
    \label{fig:fg}
  \end{subfigure}
  \hfill
  \begin{subfigure}{0.495\linewidth}
    \includegraphics[width=1\linewidth]{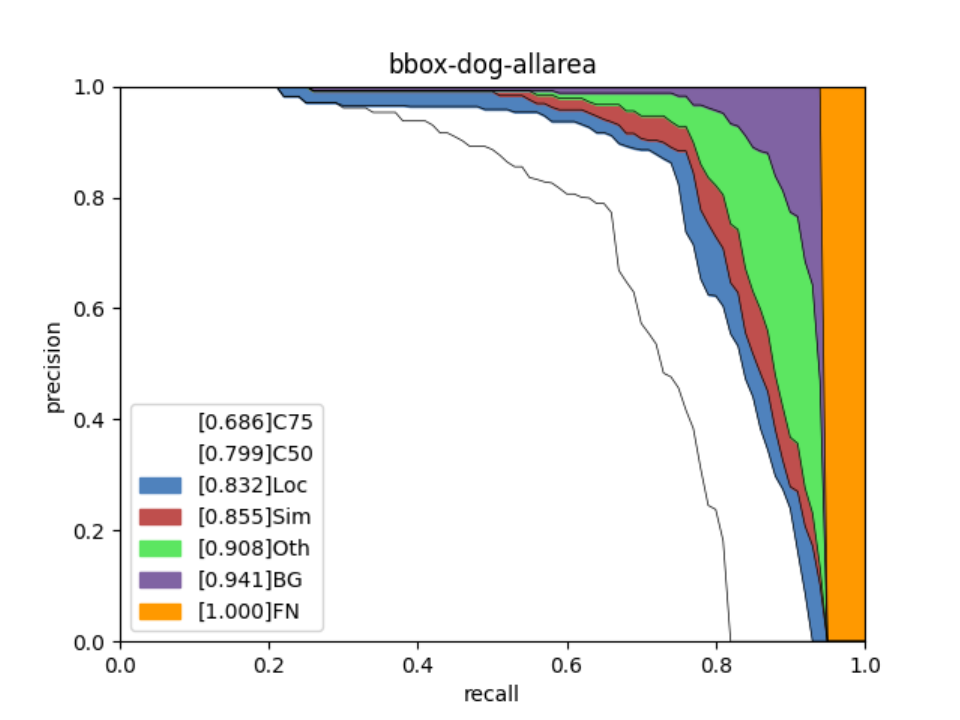}
    \caption{KD: background}
    \label{fig:bg}
  \end{subfigure}
  \hfill
  \begin{subfigure}{0.7\linewidth}
    \includegraphics[width=1\linewidth]{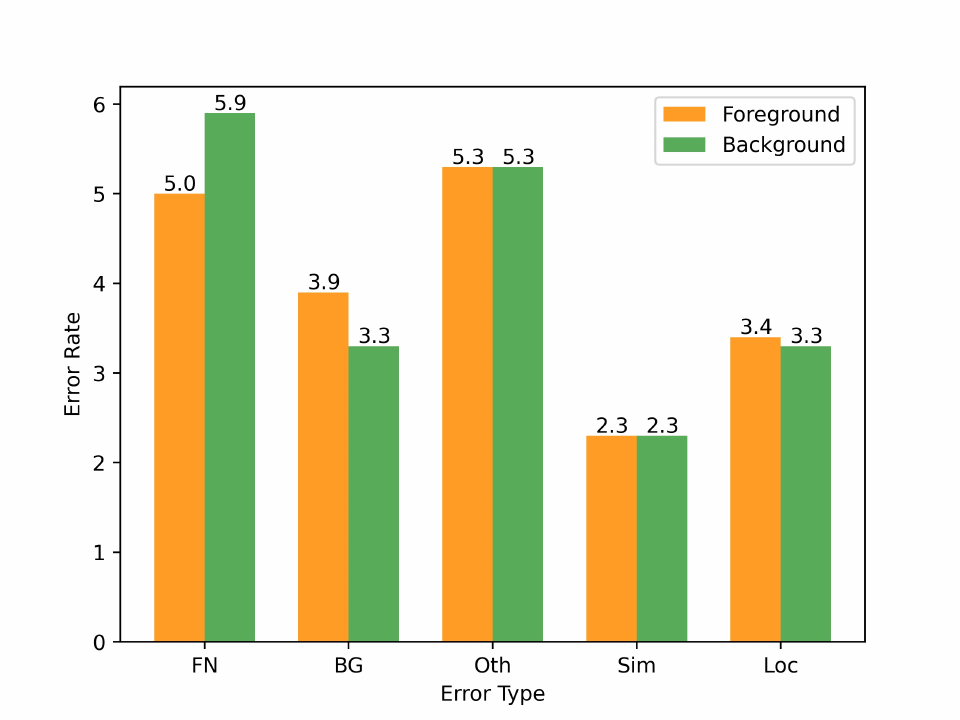}
    \caption{Error of different types analysis}
    \label{fig:fbg_com}
  \end{subfigure}
  \caption{Different error types analyses of foreground and background distillation. {\bf FN}: false negative prediction; {\bf BG}: Background false positive prediction; {\bf Oth}: classification errors; {\bf Sim}: wrong class but correct supercategory; {\bf Loc}: localization errors}
  \label{figure:fbg_error}
\end{figure}

 As shown in \cref{table:fbg ablation}, we find distilling just on foreground or background both lead significant improvements. Here we analyze different error types to investigate their effectiveness, which is shown in \cref{figure:fbg_error}. With the knowledge from background, student detectors reduce the false-positive predictions and get higher mAP. In comparison, the foreground's distillation helps students detect more targets and reduce the false-negative predictions. In conclusion, the results show that both foreground and background are crucial and have different functions for the student detectors. 

In this paper, we utilize the spatial and channel attention mask of the teacher to guide the student to focus on crucial parts. Here we conduct experiments with RetinaNet to explore the effects of each mask, which is shown in \cref{table:attention mask ablation}. Each attention mask improves the performance, especially the spatial attention mask which brings 2.6 mAP gains and 2.2 mAR gains. And the combination of two masks gets the best result. The experiments show both the attention masks help the student perform better.
\begin{table}
  \centering
  \begin{tabular}{@{}l|c|ccc}
    \toprule
    Method & \multicolumn{4}{c}{ReinaNet ResNeXt101-Res50}\\
    \midrule
    Spatial attention  & - &\checkmark&-&\checkmark\\
    Channel attention  & - &-&\checkmark&\checkmark\\
    \midrule
    mAP & 37.4 &40.0&39.7&{\bf40.2}\\
    AP$_{S}$ & 20.0 &22.3&22.0&{\bf22.8}\\
    AP$_{M}$ & 40.7 &44.0&43.5&{\bf44.0}\\
    AP$_{L}$ & 49.7 &53.6&53.4&{\bf54.0}\\
    \midrule
    mAR & 53.9 &56.1&55.8&{\bf56.2}\\
    AR$_{S}$ & 33.1 &36.5&35.7&{\bf36.8}\\
    AR$_{M}$ & 57.7 &60.2&59.9&{\bf60.3}\\
    AR$_{L}$ & 70.2 &72.1&71.8&{\bf72.3}\\
    \bottomrule
  \end{tabular}
  \caption{Ablation study of the spatial and channel attention mask.}
  \label{table:attention mask ablation}
\end{table}

\subsubsection{Sensitivity study of global distillation}
In global distillation, we rebuild the relation between different pixels to compensate for the missing global information in focal distillation and transfer it from the teacher detector to the student detector. In this subsection, we distill the student just using the global distillation with GcBlock\cite{cao2019gcnet} or Non-local module\cite{wang2018non} on Faster RCNN, which is shown in \cref{table:relation methods}. The results show both two relation methods extract effective global information and bring the student effective improvement, especially the GcBlock which brings 3.1 mAP improvement.
\begin{table}
  \centering
  \begin{tabular}{@{}c|cccc}
    \toprule
    Methods & mAP& AP$_{S}$&AP$_{M}$&AP$_{L}$\\
    \midrule
    baseline  &38.4&21.5&42.1&50.3\\
    Non-Local  &39.8 &22.7&43.1&52.3\\
    GcBlock  &{\bf41.5}&{\bf23.4}&{\bf46.0}&{\bf55.3}\\
    \bottomrule
  \end{tabular}
  \caption{Comparison of different global relation methods on Faster RCNN ResNeXt101-Res50. Here we train the student just with global distillation.}
  \label{table:relation methods}
\end{table}

\subsubsection{Sensitivity study of  $T$}
In \cref{s-a} and \cref{c-a}, we use the temperature hyper-parameter $T$ to adjust the pixels and channels distribution of the feature map. The gap between pixels and channels becomes wider and smaller when $T<1$ and $T>1$, respectively.  Here we conduct several experiments to investigate the influence of $T$. As shown in \cref{table:T ablation}, when $T=0.5$, the student gains 0.2 mAP and 0.2 mAR improvement compared with $T=1$, which means distillation without distribution adjustment. With $T=0.5$, the pixels and channels of high value are emphasized more and this helps the student detector focus on such crucial parts more and perform better. It is also observed the worst result is just a 0.4 mAP drop compared with the best result, indicating our method is not sensitive to the hyper-parameter $T$.
\begin{table}
  \centering
  \begin{tabular}{@{}c|ccccc}
    \toprule
    T & 0.3 & 0.5 & 0.8 & 1.0 & 1.2\\
    \midrule
    mAP  &40.1&{\bf40.4}&{\bf40.4}&40.2&40.0\\
    mAR  &56.4&{\bf56.7}&56.6&56.5&56.4\\
    \bottomrule
  \end{tabular}
  \caption{Ablation study of temperature hyper-parameter $T$ on RetinaNet ResNeXt101-Res50.}
  \label{table:T ablation}
\end{table}

\section{Conclusion}
\label{sec:conclusion}
In this paper, we point out the student detector needs to pay attention to both the crucial parts and global relations from the teacher. Then we propose Focal and Global Distillation (FGD) to guide the student detectors. Extensive experiments on various detectors prove that our method is simple and efficient. Furthermore, our method is just based on the feature so that FGD can be applied to two-stage detectors, anchor-based one-stage detectors, and anchor-free one-stage detectors easily. The analysis shows that the student gets a really similar feature with the teacher and initializing the student with the teacher's parameters can bring another improvement. However, our understanding of how to get a better head is preliminary and left as future works. 

{\bf Acknowledgement.} This work was supported by the NSFC project Grant No.U1833101, the SZSTI project Grant No.JCYJ20190809172201639 and Grant NO.WDZC20200820200655001.

{\small
\bibliographystyle{ieee_fullname}
\bibliography{egbib}

\begin{thebibliography}{10}\itemsep=-1pt

\bibitem{cai2018cascade}
Zhaowei Cai and Nuno Vasconcelos.
\newblock Cascade r-cnn: Delving into high quality object detection.
\newblock In {\em Proceedings of the IEEE conference on computer vision and
  pattern recognition}, pages 6154--6162, 2018.

\bibitem{cao2019gcnet}
Yue Cao, Jiarui Xu, Stephen Lin, Fangyun Wei, and Han Hu.
\newblock Gcnet: Non-local networks meet squeeze-excitation networks and
  beyond.
\newblock In {\em Proceedings of the IEEE/CVF International Conference on
  Computer Vision Workshops}, pages 0--0, 2019.

\bibitem{chen2017learning}
Guobin Chen, Wongun Choi, Xiang Yu, Tony Han, and Manmohan Chandraker.
\newblock Learning efficient object detection models with knowledge
  distillation.
\newblock {\em Advances in neural information processing systems}, 30, 2017.

\bibitem{mmdetection}
Kai Chen, Jiaqi Wang, Jiangmiao Pang, Yuhang Cao, Yu Xiong, Xiaoxiao Li,
  Shuyang Sun, Wansen Feng, Ziwei Liu, Jiarui Xu, Zheng Zhang, Dazhi Cheng,
  Chenchen Zhu, Tianheng Cheng, Qijie Zhao, Buyu Li, Xin Lu, Rui Zhu, Yue Wu,
  Jifeng Dai, Jingdong Wang, Jianping Shi, Wanli Ouyang, Chen~Change Loy, and
  Dahua Lin.
\newblock {MMDetection}: Open mmlab detection toolbox and benchmark.
\newblock {\em arXiv preprint arXiv:1906.07155}, 2019.

\bibitem{cordts2016cityscapes}
Marius Cordts, Mohamed Omran, Sebastian Ramos, Timo Rehfeld, Markus Enzweiler,
  Rodrigo Benenson, Uwe Franke, Stefan Roth, and Bernt Schiele.
\newblock The cityscapes dataset for semantic urban scene understanding.
\newblock In {\em Proceedings of the IEEE conference on computer vision and
  pattern recognition}, pages 3213--3223, 2016.

\bibitem{dai2021general}
Xing Dai, Zeren Jiang, Zhao Wu, Yiping Bao, Zhicheng Wang, Si Liu, and Erjin
  Zhou.
\newblock General instance distillation for object detection.
\newblock In {\em Proceedings of the IEEE/CVF Conference on Computer Vision and
  Pattern Recognition}, pages 7842--7851, 2021.

\bibitem{duan2019centernet}
Kaiwen Duan, Song Bai, Lingxi Xie, Honggang Qi, Qingming Huang, and Qi Tian.
\newblock Centernet: Keypoint triplets for object detection.
\newblock In {\em Proceedings of the IEEE/CVF International Conference on
  Computer Vision}, pages 6569--6578, 2019.

\bibitem{ge2021yolox}
Zheng Ge, Songtao Liu, Feng Wang, Zeming Li, and Jian Sun.
\newblock Yolox: Exceeding yolo series in 2021.
\newblock {\em arXiv preprint arXiv:2107.08430}, 2021.

\bibitem{guo2021distilling}
Jianyuan Guo, Kai Han, Yunhe Wang, Han Wu, Xinghao Chen, Chunjing Xu, and Chang
  Xu.
\newblock Distilling object detectors via decoupled features.
\newblock In {\em Proceedings of the IEEE/CVF Conference on Computer Vision and
  Pattern Recognition}, pages 2154--2164, 2021.

\bibitem{he2017mask}
Kaiming He, Georgia Gkioxari, Piotr Doll{\'a}r, and Ross Girshick.
\newblock Mask r-cnn.
\newblock In {\em Proceedings of the IEEE international conference on computer
  vision}, pages 2961--2969, 2017.

\bibitem{he2016deep}
Kaiming He, Xiangyu Zhang, Shaoqing Ren, and Jian Sun.
\newblock Deep residual learning for image recognition.
\newblock In {\em Proceedings of the IEEE conference on computer vision and
  pattern recognition}, pages 770--778, 2016.

\bibitem{heo2019comprehensive}
Byeongho Heo, Jeesoo Kim, Sangdoo Yun, Hyojin Park, Nojun Kwak, and Jin~Young
  Choi.
\newblock A comprehensive overhaul of feature distillation.
\newblock In {\em Proceedings of the IEEE/CVF International Conference on
  Computer Vision}, pages 1921--1930, 2019.

\bibitem{hinton2015distilling}
Geoffrey Hinton, Oriol Vinyals, and Jeff Dean.
\newblock Distilling the knowledge in a neural network.
\newblock {\em arXiv preprint arXiv:1503.02531}, 2015.

\bibitem{hu2018relation}
Han Hu, Jiayuan Gu, Zheng Zhang, Jifeng Dai, and Yichen Wei.
\newblock Relation networks for object detection.
\newblock In {\em Proceedings of the IEEE conference on computer vision and
  pattern recognition}, pages 3588--3597, 2018.

\bibitem{hu2018squeeze}
Jie Hu, Li Shen, and Gang Sun.
\newblock Squeeze-and-excitation networks.
\newblock In {\em Proceedings of the IEEE conference on computer vision and
  pattern recognition}, pages 7132--7141, 2018.

\bibitem{kang2021instance}
Zijian Kang, Peizhen Zhang, Xiangyu Zhang, Jian Sun, and Nanning Zheng.
\newblock Instance-conditional knowledge distillation for object detection.
\newblock {\em arXiv preprint arXiv:2110.12724}, 2021.

\bibitem{li2017mimicking}
Quanquan Li, Shengying Jin, and Junjie Yan.
\newblock Mimicking very efficient network for object detection.
\newblock In {\em Proceedings of the ieee conference on computer vision and
  pattern recognition}, pages 6356--6364, 2017.

\bibitem{li2020generalized}
Xiang Li, Wenhai Wang, Lijun Wu, Shuo Chen, Xiaolin Hu, Jun Li, Jinhui Tang,
  and Jian Yang.
\newblock Generalized focal loss: Learning qualified and distributed bounding
  boxes for dense object detection.
\newblock {\em Advances in Neural Information Processing Systems},
  33:21002--21012, 2020.

\bibitem{lin2017feature}
Tsung-Yi Lin, Piotr Doll{\'a}r, Ross Girshick, Kaiming He, Bharath Hariharan,
  and Serge Belongie.
\newblock Feature pyramid networks for object detection.
\newblock In {\em Proceedings of the IEEE conference on computer vision and
  pattern recognition}, pages 2117--2125, 2017.

\bibitem{lin2017focal}
Tsung-Yi Lin, Priya Goyal, Ross Girshick, Kaiming He, and Piotr Doll{\'a}r.
\newblock Focal loss for dense object detection.
\newblock In {\em Proceedings of the IEEE international conference on computer
  vision}, pages 2980--2988, 2017.

\bibitem{lin2014microsoft}
Tsung-Yi Lin, Michael Maire, Serge Belongie, James Hays, Pietro Perona, Deva
  Ramanan, Piotr Doll{\'a}r, and C~Lawrence Zitnick.
\newblock Microsoft coco: Common objects in context.
\newblock In {\em European conference on computer vision}, pages 740--755.
  Springer, 2014.

\bibitem{liu2016ssd}
Wei Liu, Dragomir Anguelov, Dumitru Erhan, Christian Szegedy, Scott Reed,
  Cheng-Yang Fu, and Alexander~C Berg.
\newblock Ssd: Single shot multibox detector.
\newblock In {\em European conference on computer vision}, pages 21--37.
  Springer, 2016.

\bibitem{paszke2019pytorch}
Adam Paszke, Sam Gross, Francisco Massa, Adam Lerer, James Bradbury, Gregory
  Chanan, Trevor Killeen, Zeming Lin, Natalia Gimelshein, Luca Antiga, et~al.
\newblock Pytorch: An imperative style, high-performance deep learning library.
\newblock {\em Advances in neural information processing systems},
  32:8026--8037, 2019.

\bibitem{redmon2018yolov3}
Joseph Redmon and Ali Farhadi.
\newblock Yolov3: An incremental improvement.
\newblock {\em arXiv preprint arXiv:1804.02767}, 2018.

\bibitem{ren2015faster}
Shaoqing Ren, Kaiming He, Ross Girshick, and Jian Sun.
\newblock Faster r-cnn: Towards real-time object detection with region proposal
  networks.
\newblock {\em Advances in neural information processing systems}, 28:91--99,
  2015.

\bibitem{romero2014fitnets}
Adriana Romero, Nicolas Ballas, Samira~Ebrahimi Kahou, Antoine Chassang, Carlo
  Gatta, and Yoshua Bengio.
\newblock Fitnets: Hints for thin deep nets.
\newblock {\em arXiv preprint arXiv:1412.6550}, 2014.

\bibitem{ronneberger2015u}
Olaf Ronneberger, Philipp Fischer, and Thomas Brox.
\newblock U-net: Convolutional networks for biomedical image segmentation.
\newblock In {\em International Conference on Medical image computing and
  computer-assisted intervention}, pages 234--241. Springer, 2015.

\bibitem{sun2020distilling}
Ruoyu Sun, Fuhui Tang, Xiaopeng Zhang, Hongkai Xiong, and Qi Tian.
\newblock Distilling object detectors with task adaptive regularization.
\newblock {\em arXiv preprint arXiv:2006.13108}, 2020.

\bibitem{tian2019fcos}
Zhi Tian, Chunhua Shen, Hao Chen, and Tong He.
\newblock Fcos: Fully convolutional one-stage object detection.
\newblock In {\em Proceedings of the IEEE/CVF international conference on
  computer vision}, pages 9627--9636, 2019.

\bibitem{tung2019similarity}
Frederick Tung and Greg Mori.
\newblock Similarity-preserving knowledge distillation.
\newblock In {\em Proceedings of the IEEE/CVF International Conference on
  Computer Vision}, pages 1365--1374, 2019.

\bibitem{wang2019distilling}
Tao Wang, Li Yuan, Xiaopeng Zhang, and Jiashi Feng.
\newblock Distilling object detectors with fine-grained feature imitation.
\newblock In {\em Proceedings of the IEEE/CVF Conference on Computer Vision and
  Pattern Recognition}, pages 4933--4942, 2019.

\bibitem{wang2018non}
Xiaolong Wang, Ross Girshick, Abhinav Gupta, and Kaiming He.
\newblock Non-local neural networks.
\newblock In {\em Proceedings of the IEEE conference on computer vision and
  pattern recognition}, pages 7794--7803, 2018.

\bibitem{wang2020solo}
Xinlong Wang, Tao Kong, Chunhua Shen, Yuning Jiang, and Lei Li.
\newblock Solo: Segmenting objects by locations.
\newblock In {\em European Conference on Computer Vision}, pages 649--665.
  Springer, 2020.

\bibitem{woo2018cbam}
Sanghyun Woo, Jongchan Park, Joon-Young Lee, and In~So Kweon.
\newblock Cbam: Convolutional block attention module.
\newblock In {\em Proceedings of the European conference on computer vision
  (ECCV)}, pages 3--19, 2018.

\bibitem{xie2017aggregated}
Saining Xie, Ross Girshick, Piotr Doll{\'a}r, Zhuowen Tu, and Kaiming He.
\newblock Aggregated residual transformations for deep neural networks.
\newblock In {\em Proceedings of the IEEE conference on computer vision and
  pattern recognition}, pages 1492--1500, 2017.

\bibitem{yang2019reppoints}
Ze Yang, Shaohui Liu, Han Hu, Liwei Wang, and Stephen Lin.
\newblock Reppoints: Point set representation for object detection.
\newblock In {\em Proceedings of the IEEE/CVF International Conference on
  Computer Vision}, pages 9657--9666, 2019.

\bibitem{yim2017gift}
Junho Yim, Donggyu Joo, Jihoon Bae, and Junmo Kim.
\newblock A gift from knowledge distillation: Fast optimization, network
  minimization and transfer learning.
\newblock In {\em Proceedings of the IEEE Conference on Computer Vision and
  Pattern Recognition}, pages 4133--4141, 2017.

\bibitem{zagoruyko2016paying}
Sergey Zagoruyko and Nikos Komodakis.
\newblock Paying more attention to attention: Improving the performance of
  convolutional neural networks via attention transfer.
\newblock {\em arXiv preprint arXiv:1612.03928}, 2016.

\bibitem{zhang2020improve}
Linfeng Zhang and Kaisheng Ma.
\newblock Improve object detection with feature-based knowledge distillation:
  Towards accurate and efficient detectors.
\newblock In {\em International Conference on Learning Representations}, 2020.

\end{thebibliography}
}

\clearpage
\appendix
\section*{Appendix}
\section{Experiments on recent stronger models}
We also evaluate FGD with some recent stronger models such as GFL\cite{li2020generalized}, SOLO\cite{wang2020solo} and YOLOX\cite{ge2021yolox}. The distillation settings and results are shown in \cref{table:extra results}. As it shows, FGD still can bring excellent mAP improvement for the recent stronger models including GFL, SOLO, and YOLOX.
\begin{table}
  \centering
  \begin{tabular}{l|cccc}
    \toprule
    {\bf COCO} & mAP & AP$_{S}$ & AP$_{M}$ & AP$_{L}$\\
    \midrule
    GFL-R101(T, ms) & 44.9 &28.0&49.1&57.2\\
    GFL-R50(S, 1x) & 40.2 &23.3&44.0&52.2\\
    +FGD & 43.5&26.2&47.6&56.7 \\
    \midrule
    Solo-R101(T, ms) & 37.1 &15.0&39.3&54.1\\
    Solo-R50(S, 1x) & 33.1 &12.2&36.1&50.8\\
    +FGD & 36.0&14.5&39.5&54.5\\
    \midrule
    YOLOX-L(T) & 48.5 &31.3&53.3&64.1\\
    YOLOX-M(S) & 45.1 &27.1&49.9&60.4\\
    +FGD & 46.6&29.1&51.8&61.4\\
    \bottomrule
  \end{tabular}
  \caption{Results of distillation with some recent stronger models. All the AP for Solo means Mask AP.}
 \label{table:extra results}
\end{table}

\section{Experiments on CityScapes}
Here we use FasterRCNN Res101-50 to show the effectiveness of our method on Cityscapes\cite{cordts2016cityscapes}, which is shown in \cref{table:city results}. As it shows, FGD also brings the student excellent AP improvement on CityScapes.
\begin{table}
  \centering
  \begin{tabular}{l|cccc}
    \toprule
    {\bf CityScapes} & mAP & AP$_{S}$ & AP$_{M}$ & AP$_{L}$\\
    \midrule
    RCNN-Res101(T) & 42.4 &18.2&42.9&61.8\\
    RCNN-Res50(S) & 40.2 &17.7&40.6&61.2\\
    +FGD & 43.0&18.0&42.4&61.4 \\
    \bottomrule
  \end{tabular}
\caption{Results of Faster-RCNN on CityScapes dataset.}
 \label{table:city results}
\end{table}

\section{Comparison with more methods}
Here we compare more distillation methods\cite{romero2014fitnets,li2017mimicking,heo2019comprehensive} by using Faster RCNN-Res101 to distill Faster RCNN-Res50 on COCO dataset. As \cref{table:more methods} shows, FGD also surpasses the other three methods significantly.

\begin{table}
  \centering
  \begin{tabular}{l|cccc}
    \toprule
    {\bf COCO} & mAP & AP$_{S}$ & AP$_{M}$ & AP$_{L}$\\
    \midrule
    RCNN-Res50(S) & 38.4 &21.5&42.1&50.3\\
    +FitNet\cite{romero2014fitnets} &38.9 &21.9&42.2&51.6\\
    +Mimicking\cite{li2017mimicking} &39.6 &22.5&42.8&52.2\\
    +OverHaul\cite{heo2019comprehensive} &38.9 &21.8&42.7&50.7\\
    +FGD & 40.5&22.6&44.7&53.2 \\
    \bottomrule
  \end{tabular}
\caption{Results of more distillation methods on COCO dataset.}
 \label{table:more methods}
\end{table}

\section{Sensitivity study of hyper-parameters}
\begin{figure}
  \centering
  \includegraphics[width=0.95\linewidth]{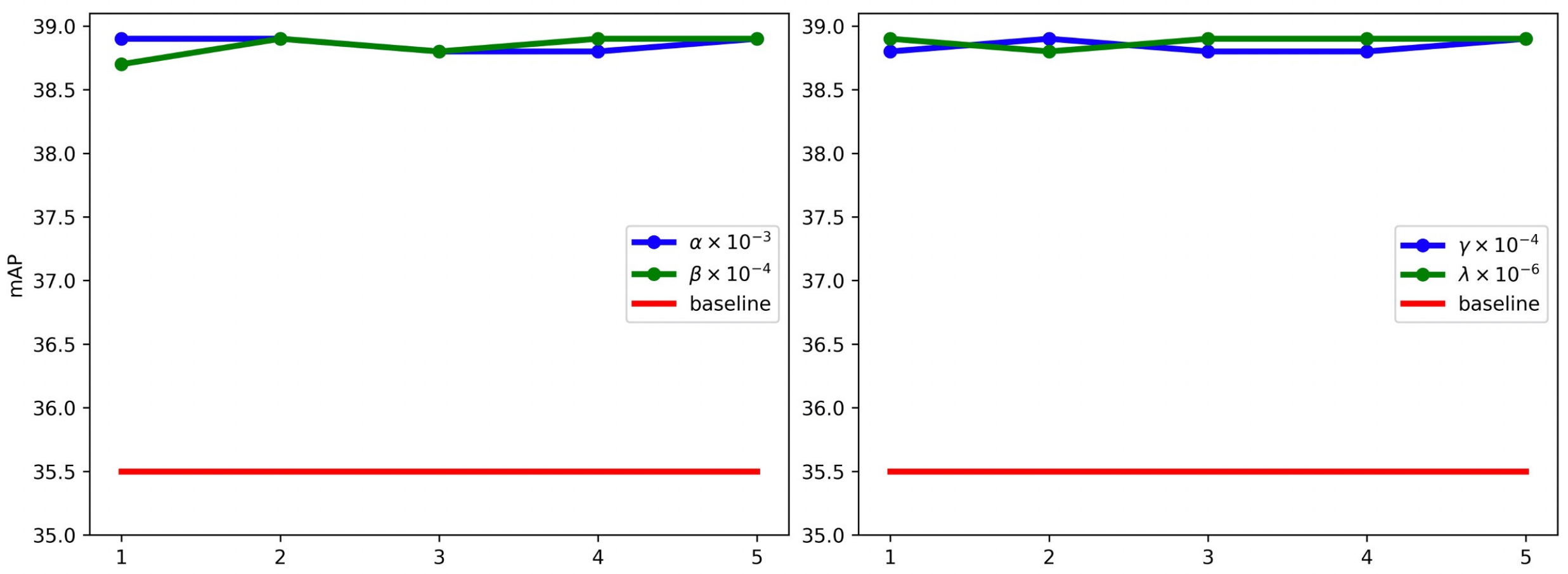}
  \caption{Hyperparameters sensitivity study of $\alpha, \beta, \gamma, \lambda$ with RetinaNet-ResX101-50(half channel) on COCO.}
  \label{figure:sens}
\end{figure}

There are five hyper-parameters in FGD. The study of $T$ is shown in \cref{table:T ablation}. The other hyper-parameters $\alpha,\beta,\gamma,\lambda$ are used for balancing the losses, which is normal in multi-loss tasks. As shown in \cref{figure:sens}, FGD is not sensitive to them.

\section{Limitations about distillation}
Knowledge distillation aims at transferring the information from the teacher to the student. So it may let student model inherit some potential biases from the teacher.

\end{document}